\documentclass{article}

\usepackage{microtype}
\usepackage{graphicx}
\usepackage{subfigure}
\usepackage{booktabs} 
\usepackage{hyperref}

\usepackage[accepted]{icml2025}

\usepackage{amsmath}
\usepackage{amssymb}
\usepackage{mathtools}
\usepackage{amsthm}
\usepackage{amscd}

\usepackage[capitalize,noabbrev]{cleveref}

\usepackage[textsize=tiny]{todonotes}

\usepackage{amsmath,amsfonts, amsthm, amssymb, bm}

\def\eqref#1{equation~\ref{#1}}

\def\1{\bm{1}}

\def\rvf{{\mathbf{f}}}

\def\rvg{{\mathbf{g}}}

\def\rvh{{\mathbf{h}}}

\def\rvv{{\mathbf{v}}}
\def\rvw{{\mathbf{w}}}

\def\rvx{{\mathbf{x}}}
\def\rvy{{\mathbf{y}}}
\def\rvz{{\mathbf{z}}}
\def\rvzero{\mathbf{0}}

\def\rmZ{{\mathbf{Z}}}

\def\evf{{f}}
\def\evg{{g}}

\def\evk{{k}}

\def\evp{{p}}

\def\evv{{v}}
\def\evw{{w}}
\def\evx{{x}}
\def\evy{{y}}

\def\mI{{\bm{I}}}

\def\msigma{{\bm{\sigma}}}
\def\mSigma{{\bm{\Sigma}}}
\def\mxi{\bm\xi}
\def\mdelta{\bm\delta}

\DeclareMathAlphabet{\mathsfit}{\encodingdefault}{\sfdefault}{m}{sl}
\SetMathAlphabet{\mathsfit}{bold}{\encodingdefault}{\sfdefault}{bx}{n}

\def\gH{{\mathcal{H}}}

\def\sF{{\mathbb{F}}}

\newcommand{\E}{\mathbb{E}}

\newcommand{\R}{\mathbb{R}}

\newcommand{\norm}[1]{||#1||}
\newcommand{\loss}{\mathcal{E}}
\theoremstyle{plain}
\newtheorem{theorem}{Theorem}[section]

\theoremstyle{definition}

\theoremstyle{remark}
\newtheorem{remark}[theorem]{Remark}
\DeclareMathOperator*{\argmin}{arg\,min}

\DeclareMathOperator{\dif}{d \!}        
\newcommand{\dt}{\dif t}

\icmltitlerunning{Learning Structured SDEs}

\makeatletter
\renewcommand{\ICML@appearing}{}  
\makeatother

\begin{document}

\twocolumn[
\icmltitle{Learning Stochastic Dynamical Systems with Structured Noise}

\icmlsetsymbol{equal}{*}

\begin{icmlauthorlist}
\icmlauthor{Ziheng Guo}{equal,uh}
\icmlauthor{James Greene}{equal,cu}
\icmlauthor{Ming Zhong}{equal,uh}
\end{icmlauthorlist}

\icmlaffiliation{uh}{Department of Mathematics, University of Houston, Texas, USA}
\icmlaffiliation{cu}{Department of Mathematics, Clarkson University, New York, USA}

\icmlcorrespondingauthor{Ming Zhong}{mzhong3@central.uh.edu}

\icmlkeywords{Nonparametric learning, stochastic dynamics, particle systems, collective dynamics}

\vskip 0.3in

]
\printAffiliationsAndNotice{\icmlEqualContribution} 



\begin{abstract}
Stochastic differential equations (SDEs) are a ubiquitous modeling framework that finds applications in physics, biology, engineering, social science, and finance.  Due to the availability of large-scale data sets, there is growing interest in learning mechanistic models from observations with stochastic noise.  In this work, we present a nonparametric framework to learn both the drift and diffusion terms in systems of SDEs where the stochastic noise is singular.  Specifically, inspired by second-order equations from classical physics, we consider systems which possess structured noise, i.e. noise with a singular covariance matrix.  We provide an algorithm for constructing estimators given trajectory data and demonstrate the effectiveness of our methods via a number of examples from physics and biology.  As the developed framework is most naturally applicable to systems possessing a high degree of dimensionality reduction (i.e. symmetry), we also apply it to the high dimensional Cucker-Smale flocking model studied in collective dynamics and show that it is able to accurately infer the low dimensional interaction kernel from particle data.
\end{abstract}

\section{Introduction}\label{sec:intro}
Many problems in science and engineering possess either inherent randomness (e.g. quantum mechanics~\cite{bera2017randomness} in physics, chromosome inheritance during meiosis~\cite{heams2014randomness}), or rather appear non-deterministic due to our inability to measure or understand intrinsic dynamics (e.g. the motion of pollen grains in water, which led to the construction of Brownian motion~\cite{einstein1906theory}, or the random walk hypothesis in the stock market~\cite{lee1992stock}); we note that the underlying source of randomness is often unknown and thus remains open to a variety of interpretations.  Nevertheless, \textit{modeling} randomness has proved to be extraordinarily effective at solving scientific problems.  The natural framework for incorporating non-determinism into dynamical systems are stochastic differential equations (SDEs).

By incorporating randomness, SDEs provide a robust framework for describing evolutionary processes with noise, and are utilized in physics, biology, chemistry, finance, as well as in many other fields.  For example, the Langevin equation incorporates both deterministic and random (thermal) forces, with the later due to microscopic collisions, and offers insight into particle dynamics on scales where random forces dominate~\cite{ebeling2008stochastic}.  Many models in biology are formulated in terms of SDEs, including stochastic Lotka-Volterra equations for describing predator-prey systems~\cite{vadillo2019comparing}, disease-transmission models~\cite{ji2014threshold}, cancer cell migration and metastasis~\cite{katsaounis2023stochastic}, genetics and mutations~\cite{dingli2011stochastic}, the flocking patterns of birds~\cite{lukeman2010inferring}, line formation~\cite{GTZ2023}, swarming and synchronization~\cite{hao2023mixed}, and the schooling of fish~\cite{gautrais2012deciphering}.  When approximating continuous-time Markov chains, SDEs naturally arise in chemical reaction networks~\cite{mozgunov2018review}.  In engineering, SDEs are utilized to study problems related to the control of multi-agent systems~\cite{ma2017consensus,wan2021cooperative}, and SDEs are at the core of mathematical finance, including option pricing and the classical Black-Scholes model~\cite{black1973pricing,hull2016options}.

Many examples of SDEs in science and engineering take a \textit{structured} representation with respect to Brownian noise.  As motivation for this representation (which is discussed more generally in Section~\ref{sec:SDEs}), we consider the dynamics of a one-dimensional particle subject to deterministic ($\rvf$) and random ($\mxi$) forces.  Newton's second law then implies that the dynamics of the position $\rvy$ of the particle are governed by the second-order SDE
\begin{align}\label{eq:SDE_second_order}
    \ddot{\rvy} &= \rvf(\rvy) + \msigma_{\rvv}(\rvy,\dot{\rvy}) \mxi(t), \quad \rvy \in \R^D,
\end{align}
where the process $\mxi$ satisfies $\langle \mxi(t) \mxi(s) \rangle = \mdelta(t-s)$ (i.e. $\mxi$ a white noise process), with position and velocity dependent velocity diffusion $\msigma_{\rvv}$~\cite{burrage2007numerical}; note that~\eqref{eq:SDE_second_order} is essentially a Langevin equation.  Converting the above to a first order system describing the position ($\rvx\coloneqq\rvy$) and velocity ($\rvv\coloneqq\dot{\rvy}$), we obtain
\begin{align}
\begin{cases}
    \dif \rvx &= \rvv \dt \\
    \dif \rvv &= \rvf(\rvx) \dt + \msigma_{\rvv}(\rvx, \rvv) \dif \rvw_{t},
\end{cases}
\label{eq:SDE_second_order_system}
\end{align}
where $\rvw_{t}$ is standard Brownian motion.  Note that the noise is fundamentally singular, in the sense that $\msigma_{\rvx}=\rvzero$, or equivalently, $\msigma= \begin{bmatrix} \rvzero & \rvzero \\ \rvzero & \msigma_{\rvv}\end{bmatrix}$.  

It is the goal of this work to study structured stochastic dynamical systems inspired by~\eqref{eq:SDE_second_order_system}, which possess singular noise; we term such systems \textit{mixed stochastic differential equations} (mSDEs).  More specifically, given trajectory data, we propose an algorithm to non-parametrically infer both the drift and diffusion in mSDEs.  Furthermore, our methods will be physics-informed, in the sense that they are adapted to any dimensionality reduction assumptions (i.e. feature maps) arising from the scientific application of interest; a classical example is collective behavior in biological or robotic populations, where the dynamics may be defined via symmetric pairwise interactions.  After introducing the framework in Sections~\ref{sec:SDEs} and~\ref{sec:learning_framework}, we demonstrate its fidelity on a number of examples, including on synthetic data generated from the stochastic van der Pol oscillator and the Cucker-Smale flocking model.   We emphasize that such Langevin-type equations (e.g.~\eqref{eq:SDE_second_order_system}) have broad applicability in the sciences~\cite{pastor1994techniques}, including cancer biology~\cite{stichel2017individual,middleton2014continuum} and collective dynamics and control more generally~\cite{lukeman2010inferring,gautrais2012deciphering,choi2022controlled}.

\subsection{Related methods}\label{subsec:related_methods}
There are a variety of methods for learning dynamical systems from data, and our methods will serve to compliment these paradigms.  Example techniques include SINDy~\cite{BPK2016}, neural ODEs~\cite{chen2019neural}, physics-informed neural networks (PINNs)~\cite{RAISSI2019686}, entropic regression~\cite{almomani2020entropic}, physics-guided deep learning~\cite{yu2024learning}, and Bayesian ODEs~\cite{tronarp2021bayesian}. However, these approaches are not adapted for high-dimensional systems.  For example, the dimension of the observed data may be prohibitively large in biological applications and collective motion; the previously discussed methods typically require sparse or low-dimensional representations of the governing systems.  Furthermore, the learning methods presented here are specially designed, in that they possess innate dimensionality-reduction capabilities, and can thus capture physically meaningful properties of the governing equations: symmetry, rotation and permutation invariances, and steady state behavior.  The methods can thus produce mechanistic interaction laws that have scientific significance, and hence provide a mechanistic counterpart to the previously discussed ``general-purpose" frameworks.  We note that the methods presented here can be considered an extension and combination of the work presented in~\cite{lu2021learning, GZC2024, GCZ2024, feng2023learning}, as the mSDEs require learning on both deterministic and stochastic drifts, which have to be handled separately.
\section{Stochastic differential equations with structured noise}\label{sec:SDEs}
We consider the following general stochastic differential equation (mSDE), 
\begin{equation}\label{eq:singular_general}
    \dif\rvz_t = \rvh(\rvz_t)\dt + \msigma(\rvz_t)\dif\rvw_t, \quad \rvz_t, \rvw_t \in \R^{D}.
\end{equation}
Here $D \ge 2$, $\rvz_t$ is a random state vector driven by the drift term $\rvh: \R^D \rightarrow \R^D$, $\msigma: \R^D \rightarrow \R^{D \times D}$ is a symmetric semi-definite covariance diffusion matrix for the standard Brownian motion $\rvw_t$.  Inspired by~\eqref{eq:SDE_second_order_system}, we assume further that $\msigma$ has a singular structure, i.e. the eigenvalues of $\msigma$, $0 = \lambda_1(\msigma) = \lambda_2(\msigma) = \cdots = \lambda_{D_x}(\msigma) < \lambda_{D_x + 1}(\msigma) \le \cdots \le \lambda_D(\msigma)$ ($D_x \ge 1$). Hence we can re-write equation~\ref{eq:singular_general} as the following mixed SDE (mSDE) system
\begin{equation}\label{eq:mix_general}
\begin{cases}
    \dif\rvx_t &= \rvf(\mxi_{f}(\rvx_t, \rvy_t))\dt, \\
    \dif\rvy_t &= \rvg(\mxi_{g}(\rvx_t, \rvy_t))\dt + \msigma^{\rvy}(\rvy_t)\dif\rvw^{\rvy}_t.
\end{cases}
\end{equation}
Here $\rvx_t \in \R^{D_x}$ and $\rvy_t \in \R^{D_y}$ are the two components of $\rvz_t$, hence $D = D_x + D_y$,  with $D_x \neq D_y$ generally.  Moreover, $\rvf: \R^{d_f} \rightarrow \R^{D_x}$ is the drift for $\rvx_t$, $\mxi_{f}: \R^{D} \rightarrow \R^{d_f}$ is the reduced feature map with $1 \le d_f \le 2$, $\rvg: \R^{d_g} \rightarrow \R^{D_y}$ is the drift for $\rvy_t$, $\mxi_{g}: \R^{D} \rightarrow \R^{d_g}$ is the reduced feature map with $1 \le d_g \le 2$, $\msigma^{\rvy}: \R^{D_y} \rightarrow \R^{D_y \times D_y}$ is a symmetric positive definite matrix, and $\rvw^{\rvy}_t$ is the standard Brownian motion.  
\begin{remark}
    When we set
    \[
    \rvz_t = \begin{bmatrix} \rvx_t \\ \rvy_t \end{bmatrix}, \quad \rvh(\rvz) = \begin{bmatrix} \rvf(\mxi_f(\rvz_t)) \\ \rvg(\mxi_g(\rvz_t)) \end{bmatrix},
    \]
    and
    \[
    \msigma(\rvz_t) = \begin{bmatrix} \rvzero_{D_x \times D_x} & \rvzero_{D_x \times D_y} \\ \rvzero_{D_y \times D_x} & \msigma^{\rvy}(\rvy_t) \end{bmatrix}, \quad \rvw_t = \begin{bmatrix} \rvzero_{D_x} \\ \rvw^{\rvy}_t \end{bmatrix},
    \]
    we obtain the original mSDE system given in~\eqref{eq:singular_general} with a singular noise structure.  If $\msigma$ does not have a diagonal structure, we can also project $\rvz_t$ onto the eigendirections of $\msigma$ which corresponds to the zero eigenvalues to obtain $\rvx_t$.  A similar statement holds for $\rvy_t$.
\end{remark}
\begin{remark}
We introduce the two feature maps $\mxi_{\rvf}$ and $\mxi_{\rvg}$ due to the common assumption that most of the high-dimensional functions live on low-dimensional manifolds, and the establishment of such feature maps builds an innate dimension reduction framework for high-dimensional learning, for example the learning framework in~\cite{FMMZ2022}.
\end{remark}
As discussed in Section~\ref{sec:intro}, models of the form of~\eqref{eq:mix_general} are ubiquitous in science and engineering.  Furthermore, many such models are realized as highly complex systems; a motivating example for this work is the phenomenon of collective dynamics, which typically consists of a large number of interacting agents.  It is generally highly nontrivial to formulate and calibrate mathematical models to describe such systems, as such models typically are formulated as nonlinear stochastic dynamical systems, which are challenging to understand both analytically and numerically.  Furthermore, in many scientific scenarios (e.g. cell tracking in biology~\cite{mavska2023cell}) observed trajectory data is available, and a fundamental goal is to infer dynamics (e.g. mechanisms of interaction in collective dynamics).  Indeed, \textit{data-driven modeling} has recently experienced a surge of interest in the machine learning community, due to its capability to effectively and efficiently learn rich mathematical structure from observations, as well as to deliver accurate predictions that can be utilized in control~
 \cite{pereira2020rise}.  Although data-driven modeling techniques have been applied to various dynamical systems, it remains a relatively unexplored approach with respect to emergent behaviors in stochastic systems.  The goal of this work is to present and numerically verify an efficient algorithm for learning both the drift ($\rvf,\rvg$) and diffusion ($\msigma^{\rvy}$) in mSDE models of the form of~\eqref{eq:mix_general}.
\section{Learning framework}\label{sec:learning_framework}
We begin by introducing the basic probability notions and notations that support the proposed learning theory.  Let $(\Omega,\sF,(\sF_{t})_{0\le t\le T},\mathbb{P})$ be a filtered probability space, for a fixed and finite time horizon $T>0$. As usual, the expectation operator with respect to $\mathbb{P}$ will be denoted by $\mathbb{E}_{\mathbb{P}}$ or simply $\mathbb{E}$. For random variables $X,Y$ we write $X\sim Y$, whenever $X,Y$ have the same distribution. We consider equation \eqref{eq:singular_general} with some given initial condition $\rvz_0\sim \mu_0$

Now given the observation data (in its continuous form), i.e. $\{(\rvx_t, \rvy_t)\}_{t \in [0, T]}$ and $\rvx_0, \rvy_0 \sim \mu^{\rvx}, \mu^{\rvy}$ respectively, we find the estimator-pair $(\hat\rvf, \hat\rvg)$ by minimizing the two distinct loss functions.  First, we find $\hat\rvf$ as an approximation to $\rvf$ from optimizing the following loss function
\begin{equation}\label{eq:f_loss}
\loss_f(\tilde\rvf) = \E\big[\frac{1}{T}\int_0^T\norm{\hat\rvf(\rvx_t, \rvy_t) - \frac{\dif\rvx_t}{\dt}}^2\big], 
\end{equation}
where $\tilde\rvf \in \gH_f$; hence $\hat\rvf = \argmin_{\tilde\rvf \in \gH_f}\loss_f(\tilde\rvf)$.  We assume the expectation, $\E$, is taken over $\rvx_0 \sim \mu_{\rvx}, \rvy_0 \sim \mu_{\rvy}$.  And for $\hat\rvg$, we use the following loss function.
\begin{equation}\label{eq:g_loss}
\begin{aligned}
\loss_g(\tilde\rvg) &= \E\Big[\frac{1}{2}\big(\int_0^T<\hat\rvg(\rvx_t, \rvy_t), \bigl(\mSigma^{\rvy}\bigr)^{-1}\hat\rvg(\rvx_t, \rvy_t)>\dt \\
&\quad - 2\int_0^T<\hat\rvg(\rvx_t, \rvy_t), \bigl(\mSigma^{\rvy}\bigr)^{-1}\dif\rvy_t>\big)\Big].
\end{aligned}
\end{equation}
\begin{algorithm}[h!]
  \caption{Estimation of the Diffusion Function \(\mSigma^{\rvy}(\cdot)\) and \(\sigma^{\rvy}(\cdot)\)   from Discrete Data}
  \label{alg:Sigma_estimation}
\begin{algorithmic}
  \STATE {\bfseries Input:} 
  \STATE \quad -- Discrete observations 
             \(\bigl\{\mathbf{y}_{l}^{(m)}\bigr\}_{l=1,\ldots,L}^{m=1,\ldots,M}\) 
             with time points \(0 = t_1 < t_2 < \cdots < t_L = T\).
  \STATE \quad -- A candidate function class \(\mathcal{H}_{\mSigma^{\rvy}}\) 
             for \(\tilde{\mSigma^{\rvy}}:\mathbb{R}^d\to\mathbb{R}^{d\times d}\).
  \STATE \quad -- (Optional) A stopping criterion (e.g.\ max iterations, tolerance).

  \STATE \textbf{Step 1: Compute empirical quadratic variations.}
  \FOR{$m = 1$ {\bfseries to} $M$}
    \STATE \(\displaystyle Q^{(m)} \leftarrow \mathbf{0}_{d\times d}\).
    \FOR{$l = 1$ {\bfseries to} $L-1$}
      \STATE \(\Delta \mathbf{y}^{(m)}_l \leftarrow \mathbf{y}^{(m)}_{l+1} - \mathbf{y}^{(m)}_{l}\).
      \STATE \(Q^{(m)} \leftarrow Q^{(m)} \;+\;
             \Delta \mathbf{y}^{(m)}_l \,
             \bigl(\Delta \mathbf{y}^{(m)}_l\bigr)^\top.\)
    \ENDFOR
  \ENDFOR

  \STATE \textbf{Step 2: Define the discrete loss function.}
  \FOR{$m = 1$ {\bfseries to} $M$}
    \STATE \( I^{(m)}(\tilde{\mSigma}^{\rvy}) \leftarrow \mathbf{0}_{d\times d}\).
    \FOR{$l = 1$ {\bfseries to} $L-1$}
      \STATE \(\Delta t_l \leftarrow t_{l+1} - t_l\).
      \STATE \(I^{(m)}(\tilde{\mSigma}^{\rvy}) \leftarrow 
              I^{(m)}(\tilde{\mSigma}^{\rvy}) \;+\;
              \tilde{\mSigma}^{\rvy}\bigl(\mathbf{y}^{(m)}_l\bigr)\,\Delta t_l.\)
    \ENDFOR
  \ENDFOR
  \STATE Define 
  \[
     \mathcal{E}(\tilde{\mSigma}^{\rvy})
     \;=\;
     \frac{1}{M}
     \sum_{m=1}^{M}
       \bigl\| Q^{(m)} - I^{(m)}(\tilde{\mSigma}^{\rvy}) \bigr\|^2.
  \]

  \STATE \textbf{Step 3: Solve the minimization problem.}
  \STATE \quad\quad \(\displaystyle \hat{\mSigma}^{\rvy} \;\leftarrow\;
            \arg\min_{\tilde{\mSigma}^{\rvy} \in \mathcal{H}_{\mSigma^{\rvy}}} \,
            \mathcal{E}(\tilde{\mSigma}^{\rvy})\)
         \quad (using a suitable optimizer subject to PD constraints).

  \STATE \textbf{Step 4: Recover the diffusion coefficient \(\hat{\sigma}^{\rvy}(\cdot)\).}
   \STATE \quad In practice, we use the spectrum decomposition:
   \[
     \hat{\mSigma}^{\rvy}
     \;=\; 
     U\,D\,U^\top,
   \]
   where \(U\) is an orthonormal matrix of eigenvectors, and \(D\) is a diagonal matrix of eigenvalues (all positive).  Then set
   \[
     \hat{\sigma}^{\rvy}
     \;=\;
     U\,\sqrt{D}\,U^\top.
   \]

  \STATE {\bfseries Output:} 
  \STATE \quad (1) The estimated diffusion covariance \(\hat{\mSigma}^{\rvy}(\cdot)\).
  \STATE \quad (2) Optionally, the diffusion coefficient 
                  \(\hat{\sigma}^{\rvy}(\cdot)\).
\end{algorithmic}
\end{algorithm}
Here $\mSigma^{\rvy} = \msigma^{\rvy}(\msigma^{\rvy})^\top$ and $\tilde\rvg \in \gH_g$.  Similarly, $\hat\rvg = \argmin_{\tilde\rvg \in \gH_g}\loss_g(\tilde\rvg)$.  Both functional spaces $\gH_f$ and $\gH_g$ are chosen to be convex and compact, and the loss functions are convex, the two minimization problems have unique minimizers when optimized over $\gH_f$ and $\gH_g$.  For the details of actual implementation, see the algorithm in~\cite{lu2021learning, GCZ2024}.

We estimate the covariance diffusion matrix $\mSigma^{\rvy}$ where $\mSigma^{\rvy} = \msigma^{\rvy}(\msigma^{\rvy})^\top$ by usual quadratic (co)variation arguments. Namely, the estimation of $\mSigma^{\rvy}$ is the minimizer of the following loss function 
\begin{equation}\label{eq:sigma_loss}
        \mathcal{E}_{\Sigma}(\tilde \mSigma^{\rvy}) = \E\Big[ [\rvy, \rvy]_T - \int_{t=0}^T \tilde \Sigma^{\rvy}(\rvy_t)\dif t)\Big]^2.
    \end{equation}
where $[\rvy, \rvy]_T$ is the quadratic variation of the stochastic process $\rvy_t$ over time interval $[0, T]$.
\subsection{Algorithm for estimating diffusion}\label{subsec:algorithm}
Algorithm~\ref{alg:Sigma_estimation} shows the details on how to obtain the diffusion term.
\subsection{Performance Measures}\label{subsec:perform}
In order to properly gauge the accuracy of our learning estimators, we provide three different performance measures of our estimated drift.  First, if we have access to original drift function $\rvh$, then we will use the following error to compute the difference between $\hat\rvh$ (our estimator) to $\rvh$ with the following norm
\begin{equation}\label{eq:rho_norm}
    \norm{\rvh - \hat\rvh}_{L^2(\rho)}^2 = \int_{\R^d}\norm{\rvh(\rvz) - \hat\rvh(\rvz)}_{\ell^2(\R^d)}^2 \, \dif \rho(\rvz),
\end{equation}
where the weighted measure $\rho$,  defined on $\R^d$, is given as follows
\begin{equation}\label{eq:rho_def}
\rho(\rvz) = \E\Big[\frac{1}{T}\int_{t = 0}^T\delta_{\rvz_t}(\rvz)\Big], \quad \text{where $\rvz_t$ evolves from $\rvz_0$}
\end{equation}
and  
\[
\rvh =
\begin{bmatrix}
\rvf \\[6pt]
\rvg
\end{bmatrix},
\quad
\hat{\rvh} =
\begin{bmatrix}
\hat{\rvf} \\[6pt]
\hat{\rvg}
\end{bmatrix}.
\]
The norm given by \eqref{eq:rho_norm} is useful only from the theoretical perspective, e.g. showing convergence.  Under normal circumstances, $\rvh$ is most likely non-accessible.  Thus we look at a performance measure that compares the difference between $\{\rvz_t\}_{t \in [0, T]}$ (the observed trajectory that evolves from $\rvz_0 \sim \mu_0$ with the unknown $\rvh$) and $\{\hat\rvz_t\}_{t \in [0, T]}$ (the estimated trajectory that evolves from the same $\rvz_0$ with the learned $\hat\rvh$ and driven by the same realized random noise as used by the original dynamics).  Then, the difference between the two trajectories is measured as follows
\begin{equation}\label{eq:traj_norm}
\norm{\rmZ - \hat\rmZ} = \E\Big[\frac{1}{T}\int_{t = 0}^T\norm{\rvz_t - \hat\rvz_t}_{\ell^2(\R^d)}^2 \, \dif t\Big].
\end{equation}
However, comparing two sets of trajectories (even with the same initial condition) on the same random noise is not realistic. We compare the distribution of the trajectories over different initial conditions and all possible noise at some chosen time snapshots using the Wasserstein distance at any given time $t \in [0, T]$. Let $\mu^M_t$ be the empirical distribution at time $t$ for the simulation under $\rvh$ with $M$ trajectories, and $\hat{\mu}^M_t$ be the empirical distribution at time $t$ for the simulation with $M$ trajectories under $\hat{\rvh}$ where:
\begin{equation}
    \mu^M_t = \frac{1}{M} \sum_{i=1}^M \delta_{\rvz^{(i)}(t)}, \quad \hat{\mu}^M_t = \frac{1}{M} \sum_{i=1}^M \delta_{\hat{\rvz}^{(i)}(t)}
\end{equation}
Then the Wasserstein distance of order two between $\mu^M_t$ and $\hat{\mu}^M_t$ is calculated as
\begin{align}\label{eq:wass_dis}
\begin{split}
    &\mathcal{W}_2(\mu^M_t, \hat{\mu}^M_t \,|\, \mu_0) \\
    &= \left( \inf_{\pi \in \Pi(\mu^M_t, \hat{\mu}^M_t \,|\, \mu_0)} \int_{\R^d \times \R^d} \|x - y\|^2 \, \dif \pi(x, y) \right)^{1/2}.
\end{split}
\end{align}
Here, $\Pi(\mu^M_t, \hat{\mu}^M_t \,|\, \mu_0)$ is the set of all joint distributions on $\R^d \times \R^d$ with marginals $\mu^M_t$ and $\hat{\mu}^M_t$, and with the additional constraint that the joint distribution must be consistent with the initial distribution of $\rvz_0 \sim\mu_0$.
\section{Applications in science and engineering}\label{sec:applications}
We test our learning theory developed in Section~\ref{sec:learning_framework} on a number of synthetic data sets. We begin by considering a toy model to demonstrate that our methods are able to infer the drift and diffusion of mSDEs.  We then apply our methods to well known mSDE systems in physics and biology, including the Van der Pol oscillator, a simplified Vicsek model for active matter, the H\'{e}non-Heiles Hamiltonian system, and lastly the well-known (and high-dimensional) Cucker-Smale flocking model. Our function estimation job is carried out in basis method with \eqref{eq:f_loss}, \eqref{eq:g_loss} and \eqref{eq:sigma_loss}. The observations, serving as the input dataset for testing our method, are generated by the Euler-Maruyama scheme utilizing the drift functions as we just mentioned. The basis space $\gH$ is constructed via either B-splines or piecewise polynomials with trigonometric functions with a maximum degree ($p_{\max}$) of $2$. For systems of dimension $D \geq 2$, each basis function is derived through a tensor grid product, utilizing one-dimensional basis defined by knots that segment the domain in each dimension.  

The common parameters for the examples presented in this section are provided in Table~\ref{tab:param}; other model-specific parameters will be specified in each respective subsection. The estimation results are evaluated using several different metrics. We record the noise terms, $\dif \rvw^\rvy_t$, from the trajectory generation process and compare the trajectories produced by the estimated drift functions, $\hat{\rvg}$, under identical noise conditions. We examine trajectory-wise errors using~\eqref{eq:traj_norm} with relative trajectory error. And we calculate the relative $L^2$ error using~\eqref{eq:rho_norm}, where $\rho$ is defined by~\eqref{eq:rho_def}. Furthermore, we assess the distribution-wise discrepancies between observed and estimated results, computing the Wasserstein distance at various time steps via~\eqref{eq:wass_dis}. 
\begin{table}
    \centering
\caption{Parameter values}
\label{tab:param}
    \begin{tabular}{|c|c|} \hline 
         $T$&  1\\ \hline 
         $\Delta t$&  0.001\\ \hline 
         $M$&  3000\\ \hline
 $\mu_0$& Uniform(0,1)\\\hline
    \end{tabular}
    
\end{table}
\subsection{A toy model}\label{subsec:toy_model}
We begin with a toy model to test our learning theory. Consider the mSDE system 
\begin{equation}\label{eq:toy model}
\begin{cases}
    \dif \evx_t &= \bigl(0.4 \evx_t - 0.1 \evx_t \evy_t\bigr) \dif t, \\[6pt]
    \dif \evy_t &= \bigl(-0.8 \evy_t + 0.2 \evx_t^2\bigr) \dif t 
                  \;+\; \sigma \dif \evw^{\evy}_t.
\end{cases}
\end{equation}
Comparing with~\eqref{eq:mix_general}, we see that $\mxi_f$ and $\mxi_g$ are the identity mappings and 
\begin{align*}
\evf(\evx, \evy) &= 0.4\evx - 0.1\evx\evy \\
\evg(\evx, \evy) &= -0.8\evy + 0.2\evx^2
\end{align*} 
with $\msigma^\evy(\evy) = \sigma$.

\begin{figure}[ht]
\vskip 0.2in
\begin{center}
\centerline{\includegraphics[width=\columnwidth]{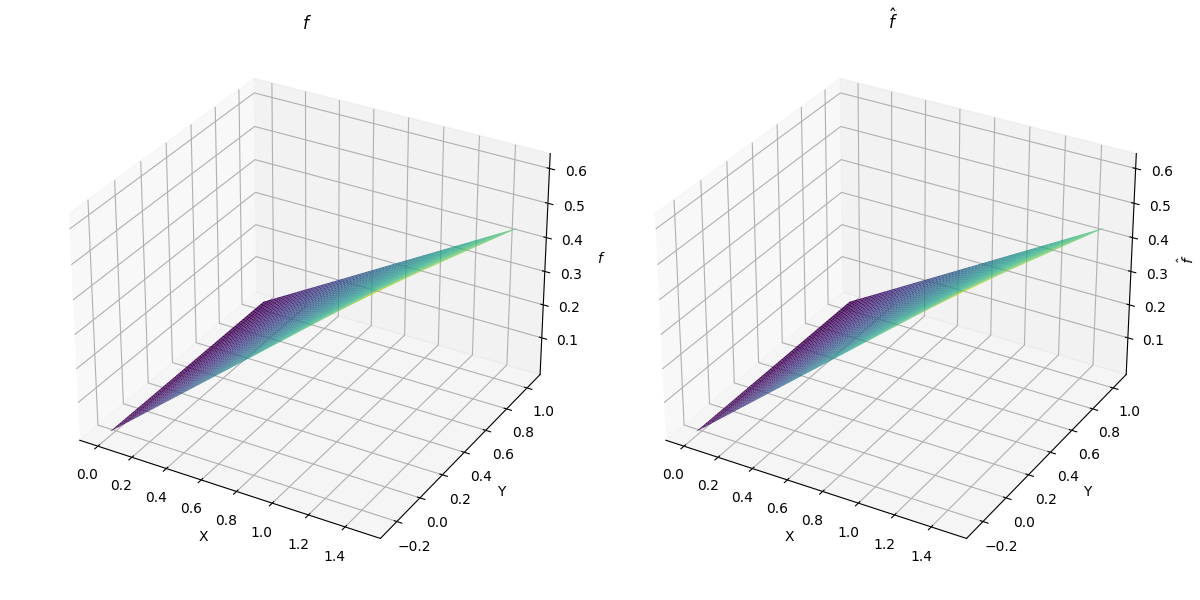}}
\caption{Comparison of $\evf$ (left) and $\hat{\evf}$ (right) for the toy model~\eqref{eq:toy model}.}
\label{fig:toy model f}
\end{center}
\vskip -0.2in
\end{figure}

\begin{figure}[ht]
\vskip 0.2in
\begin{center}
\centerline{\includegraphics[width=\columnwidth]{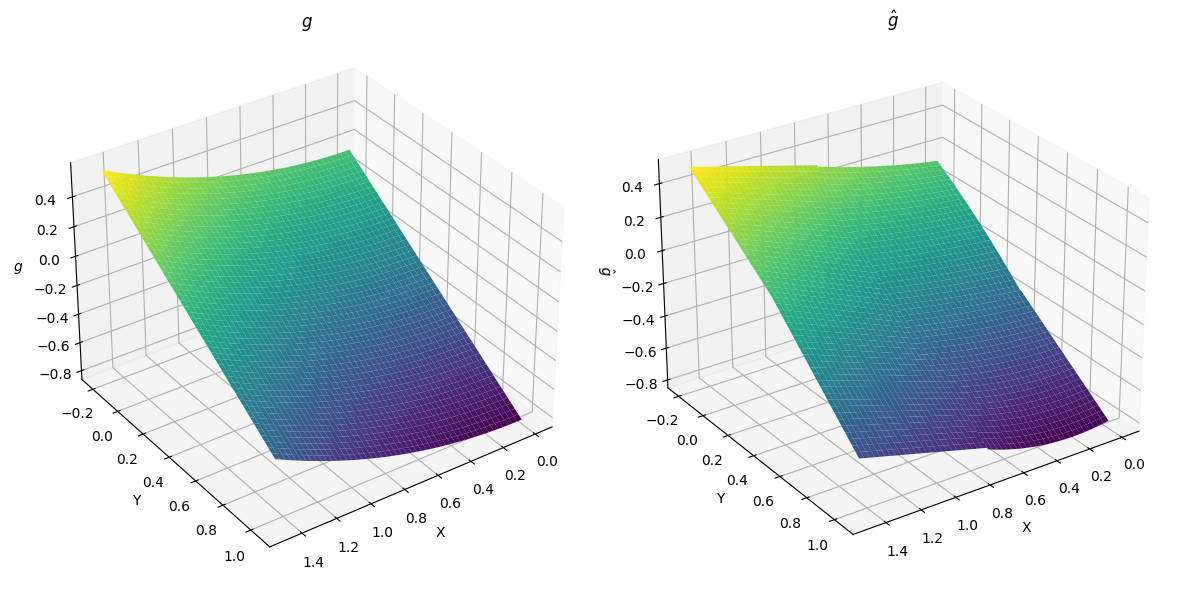}}
\caption{Comparison of $\evg$ (left) and $\hat{\evg}$ (right) for toy model}
\label{fig:toy model g}
\end{center}
\vskip -0.2in
\end{figure}

Figure~\ref{fig:toy model f} and \ref{fig:toy model g} presents the comparison of true drift function $\evf$ and $\evg$ with estimated drift function $\hat{\evf}$ and $\hat{\evg}$ respectively. Table~\ref{tab:toy model} describes the performance measures drift function estimation, while Table~\ref{tab:toy model sigma} shows our estimation result of the diffusion term. 

\begin{table}
    \centering
\caption{Toy model drift estimation summary}
\label{tab:toy model}
    \begin{tabular}{|c|c|} \hline 
         Relative $L^2(\rho)$ Error& 0.017\\ \hline 
         Relative Trajectory Error& $4.2e-3$ $\pm$ $8.6e-3$\\ \hline 
         Wasserstein Distance at $t = 25$& 0.0144\\ \hline
 Wasserstein Distance at $t = 50$&0.0149\\\hline
 Wasserstein Distance at $t = 100$& 0.0151\\\hline
    \end{tabular}
\end{table}

\begin{table}
    \centering
\caption{Toy model diffusion estimation}
\label{tab:toy model sigma}
    \begin{tabular}{|c|c|} \hline 
         True $\sigma$& Estimated $\hat{\sigma}$\\ \hline 
         0.1000& 0.1000\\ \hline
    \end{tabular}
    
\end{table}
\subsection{van der Pol oscillator}\label{subsec:one_one}
The van der Pol oscillator is a classical example of a self-sustained oscillator with nonlinear damping, which has many applications in biology and physics, including describing the action potentials of neurons~\cite{fitzhugh1961impulses,nagumo1962active}, and the rhythm synchronization of the heartbeat~\cite{dos2004rhythm}.  A two-dimensional representation of this system with Brownian noise takes the form  
\begin{equation}\label{eq:Van der Pol}
\begin{cases}
    \dif\evx_t &= \evy_t\dif t, \\[6pt]
    \dif\evy_t &= \Bigl(\mu\bigl(1 - \evx_t^2\bigr)\,\evy_t - \evx_t\Bigr)\dif t + \sigma \dif\evw_t^{\evy}
\end{cases}
\end{equation}
where \(\evx\) and \(\evy\) are state variables, and \(\mu\) is a parameter controlling the nonlinear damping.  For \(\mu > 0\), the system displays a stable limit cycle whose amplitude is regulated by the \(1 - \evx^2\) term.  We note the above van der Pol system is a specific example of the more general Li{\'e}nard systems, which have been utilized to study many phenomena in biology, including predator-prey systems, as well as chemical reaction networks~\cite{forest2007lienard}.  Li{\'e}nard systems also generally possess the form of~\eqref{eq:mix_general}, and hence this entire class of systems can be considered as a specific instance of mSDEs.

Comparing~\eqref{eq:Van der Pol} with the general framework presented in~\eqref{eq:mix_general}, we see that $\mxi_f$ and $\mxi_g$ are the identity mappings, and
\begin{align*}
\evf(\evx, \evy) &= \evy \\
\evg(\evx, \evy) &= \mu\,\bigl(1 - \evx^2\bigr)\,\evy \;-\; \evx
\end{align*}
with $\msigma^\evy(\evy) = \sigma$.

\begin{figure}[ht]
\vskip 0.2in
\begin{center}
\centerline{\includegraphics[width=\columnwidth]{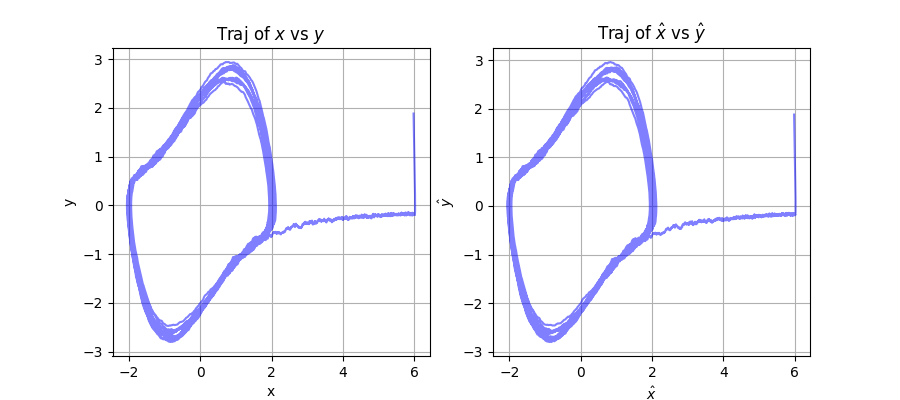}}
\caption{Van der Pol trajectory in the x-y plane. Left: Trajectory generated using the true drift function. Right: Trajectory generated using the estimated drift function.}
\label{fig:Van der Pol traj}
\end{center}
\vskip -0.2in
\end{figure}

The parameters used in our simulation to generate the synthetic trajectory data are $\mu = 1$ and $\sigma = 0.1$. To clearly observe the nonlinear limit cycle, we set $T = 100$. Figure~\ref{fig:Van der Pol traj} shows the trajectory-wise comparison, with the left providing a realization from the true dynamics $(\evx_t,\evy_t)$, while on the right we observe the corresponding estimated trajectory $(\hat{\evx}_t,\hat{\evy}_t)$, which is obtained by solving the mSDE with the estimated drifts ($\hat{\evf}$ and $\hat{\evg}$).

\begin{figure}[ht]
\vskip 0.2in
\begin{center}
\centerline{\includegraphics[width=\columnwidth]{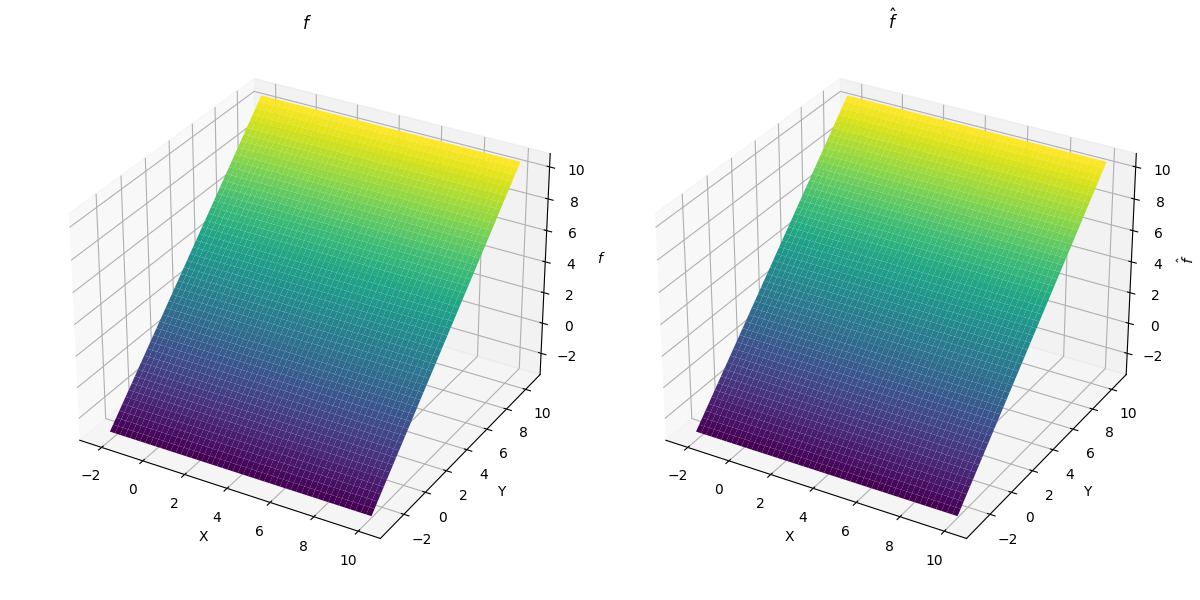}}
\caption{Comparison of $\evf$ (left) and $\hat{\evf}$ (right) for the Van der Pol oscillator~\eqref{eq:Van der Pol}.}
\label{fig:Van der Pol f}
\end{center}
\vskip -0.2in
\end{figure}

\begin{figure}[ht]
\vskip 0.2in
\begin{center}
\centerline{\includegraphics[width=\columnwidth]{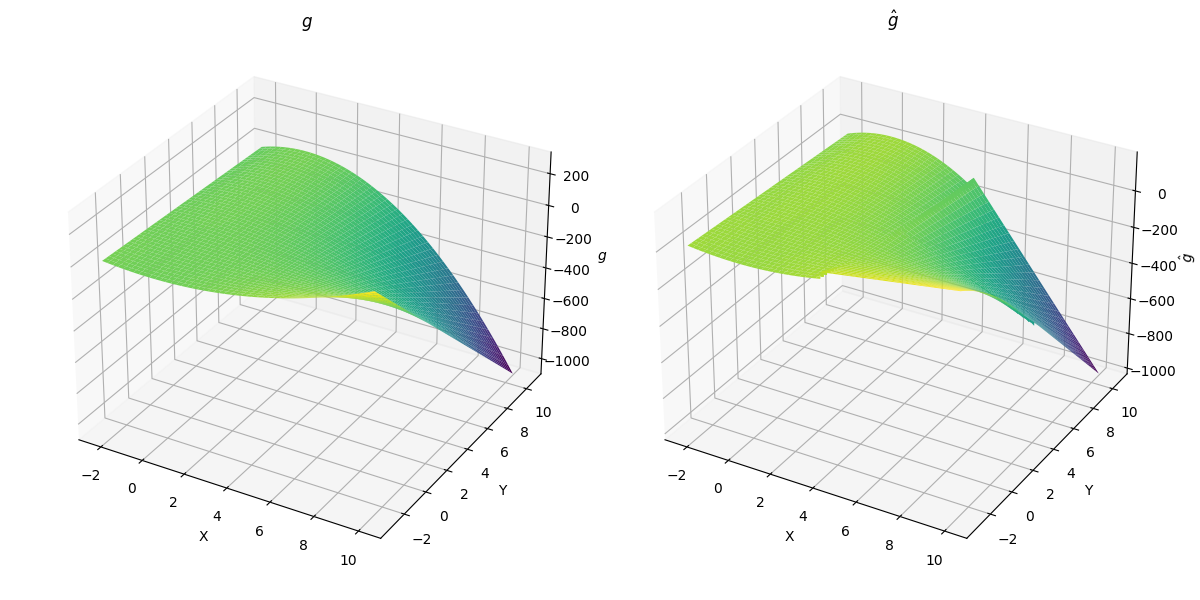}}
\caption{Comparison of $\evg$ (left) and $\hat{\evg}$ (right) for Van der Pol oscillator~\eqref{eq:Van der Pol}.}
\label{fig:Van der Pol g}
\end{center}
\vskip -0.2in
\end{figure}

Figure \ref{fig:Van der Pol f} and Figure \ref{fig:Van der Pol g} show the comparison of true drift function $\evf$ and $\evg$ with estimated drift function $\hat{\evf}$ and $\hat{\evg}$ respectively. 

Table \ref{tab:Van der Pol} summarizes the drift estimation performance for the Van der Pol oscillator. It reports the performance measures mentioned in Section~\ref{subsec:perform}, including the relative \(L^2(\rho)\) error, the relative trajectory error, and the Wasserstein distances at different time points. Table \ref{tab:Van der Pol sigma
} presents the diffusion estimation results, which compares the true noise coefficient 
\(\sigma\) with its estimated value \(\hat{\sigma}\), demonstrating a highly accurate estimation.

\begin{table}
    \centering
\caption{Van der Pol oscillator drift estimation summary}
\label{tab:Van der Pol}
    \begin{tabular}{|c|c|} \hline 
         Relative $L^2(\rho)$ Error& 0.0297\\ \hline 
         Relative Trajectory Error& $0.019$ $\pm$ $0.071$\\ \hline 
         Wasserstein Distance at $t = 25$& 0.0521\\ \hline
 Wasserstein Distance at $t = 50$&0.0548\\\hline
 Wasserstein Distance at $t = 100$& 0.0539\\\hline
    \end{tabular}
\end{table}

\begin{table}
    \centering
\caption{Van der Pol oscillator diffusion estimation}
\label{tab:Van der Pol sigma
}
    \begin{tabular}{|c|c|} \hline 
         True $\sigma$& Estimated $\hat{\sigma}$\\ \hline 
         0.1000& 0.1007\\ \hline
    \end{tabular}
    
\end{table}

\subsection{Vicsek model}\label{subsec:one_two}
The Vicsek model is a classic example of self-organized collective motion, where particles (active matter) move with constant speed and adjust their heading based on local interactions with the goal of alignment, and are subject to random (Brownian) noise~\cite{vicsek1995novel}.  Here we consider a simplified single-agent Vicsek system:
\begin{equation}\label{eq:Vicsek}
\begin{cases}
    \dif \evx_t &= \evv \cos(\theta_t)\,\dt, \\[6pt]
    \dif \evy_t &= \evv \sin(\theta_t)\,\dt, \\[6pt]
    \dif \theta_t &= \evk\bigl(\evx_t - \evy_t\bigr)\,\dt 
                     \;+\; \sigma\,\dif \evw^\theta
\end{cases}
\end{equation}
where \(\evx_t\) and \(\evy_t\) denote the agent's position, \(\theta_t\) is its orientation, 
\(\evv\) is a constant speed, \(\evk\) is an interaction parameter, 
and \(\sigma\,\dif \evw^\theta_t\) represents stochastic noise.

In this case, $\mxi_f$ and $\mxi_g$ are identity mappings and \[
\rvf(\theta) 
= \begin{bmatrix}
\evf_1(\theta)\\[6pt]
\evf_2(\theta)
\end{bmatrix}
= \begin{bmatrix}
\evv\,\cos\bigl(\theta\bigr)\\[6pt]
\evv\,\sin\bigl(\theta\bigr)
\end{bmatrix}
\]
and 
\[
\evg(\evx, \evy) \;=\; \evk\,\bigl(\evx - \evy\bigr)
\] with $\msigma^\rvy(\rvy) = \sigma$. The parameters used in our simulation are $\evv = 0.03$ and $\evk = 0.05$ and $\sigma = 0.08$. The initial distribution for the model is a uniformly distributed angle in $[0,2\pi).$  

\begin{figure}[ht]
\vskip 0.2in
\begin{center}
\centerline{\includegraphics[width=\columnwidth]{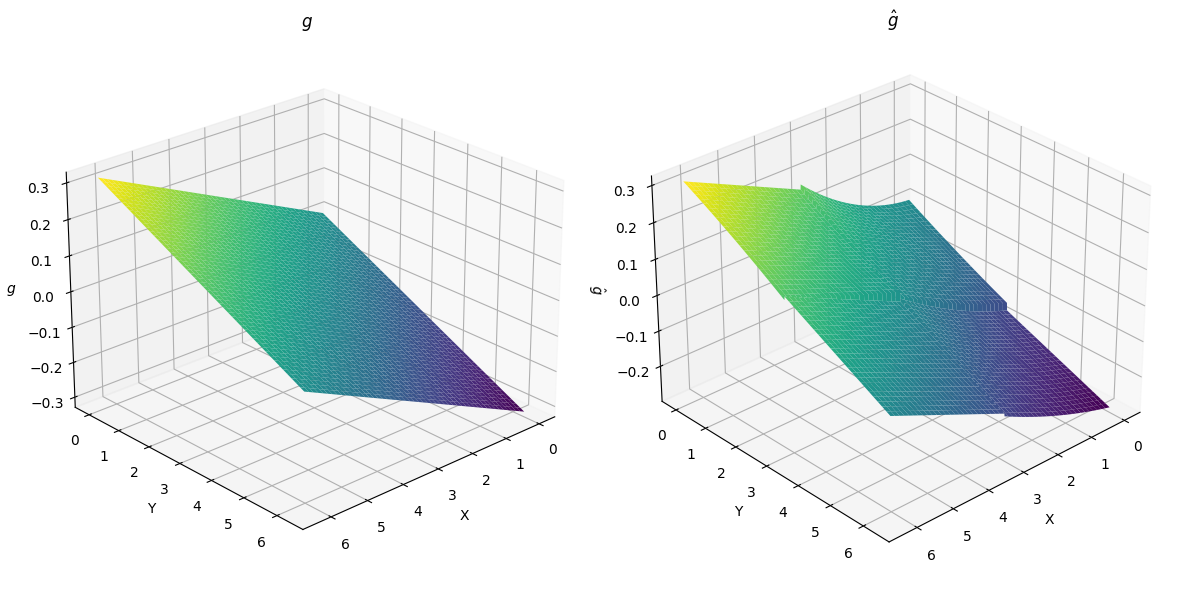}}
\caption{Comparison of $\evg$ (left) and $\hat{\evg}$ (right) for the modified Vicsek Model~\eqref{eq:Vicsek}.}
\label{fig:Vicsek g}
\end{center}
\vskip -0.2in
\end{figure}

Figure \ref{fig:Vicsek g} shows the comparison of drift function $\evg$ and estimated $\hat{\evg}$. The performance measures are displayed in Table \ref{tab:Vicsek}. The estimation result of diffusion function is shown in Table \ref{tab:Vicsek sigma}.

\begin{table}
    \centering
\caption{Modified Vicsek model drift estimation summary}
\label{tab:Vicsek}
    \begin{tabular}{|c|c|} \hline 
         Relative $L^2(\rho)$ Error& 0.044\\ \hline 
         Relative Trajectory Error& $4.63e-3$ $\pm$ $4.65e-3$\\ \hline 
         Wasserstein Distance at $t = 0.25$& 0.001075\\ \hline
 Wasserstein Distance at $t = 0.5$&0.002169\\\hline
 Wasserstein Distance at $t = 1$&0.004433\\\hline
    \end{tabular}

\end{table}

\begin{table}
    \centering
\caption{Modified Vicsek model diffusion estimation}
\label{tab:Vicsek sigma}
    \begin{tabular}{|c|c|} \hline 
         True $\sigma$& Estimated $\hat{\sigma}$\\ \hline 
         0.0800& 0.0800\\ \hline
    \end{tabular}

\end{table}
\subsection{H\'{e}non-Heiles system}\label{subsec:two_two}
The H\'{e}non-Heiles system is a classical two-degree-of-freedom Hamiltonian model, originally introduced to study chaotic motion in astronomical systems~\cite{feit1984wave}. 
A stochastic variant of this system is given by the following mSDE system:
\begin{equation}\label{eq:henon_heiles_system}
\begin{cases}
    \dif \evx_t &= \evp_{\evx,t}\,\dt, \\[6pt]
    \dif \evy_t &= \evp_{\evy,t}\,\dt, \\[6pt]
    \dif \evp_{\evx,t} &= \bigl(-\,\evx_t \;-\; 2\,\lambda\,\evx_t\,\evy_t \bigr)\,\dt 
                     \;+\; \sigma_1 \,\dif \evw^{\evp_x}_t, \\[6pt]
    \dif \evp_{\evy,t} &= \bigl(-\,\evy_t \;-\; \lambda\,(\evx_t^2 \;-\; \evy_t^2)\bigr)\,\dt 
                     \;+\; \sigma_2 \,\dif \evw^{\evp_y}_t.
\end{cases}
\end{equation}
where \((\evx,\,\evy)\) are position coordinates, \((\evp_x,\,\evp_y)\) are the associated momentum, 
\(\lambda\) is a real parameter, and \(\sigma_1\), \(\sigma_2\) are diffusion terms. 
For \(\sigma_1 = \sigma_2 = 0\), the system reduces to the original deterministic H\'{e}non-Heiles model.

In this case, compared with general model \eqref{eq:mix_general}, $\mxi_f$ and $\mxi_g$ are the identity mappings and 
\[
\rvf(\evp_{x}, \evp_{y}) 
= \begin{bmatrix}
\evf_{1}(\evp_{x}, \evp_{y}) \\[6pt]
\evf_{2}(\evp_{x}, \evp_{y})
\end{bmatrix}
= \begin{bmatrix}
\evp_{x} \\[6pt]
\evp_{y}
\end{bmatrix}
\] and 
\[
\rvg(\rvx, \rvy)
= \begin{bmatrix}
\evg_{1}(\evx, \evy)\\[6pt]
\evg_{2}(\evx, \evy)
\end{bmatrix}
= \begin{bmatrix}
-\,\evx \;-\; 2\,\lambda\,\evx\,\evy\\[6pt]
-\, \evy \;-\; \lambda\bigl(\evx^{2} - \evy^{2}\bigr)
\end{bmatrix}.
\] 
The noise structure now becomes \[
\msigma^\evy(\evy)
= \begin{bmatrix}
\sigma_{1} & 0\\[6pt]
0 & \sigma_{2}
\end{bmatrix}.
\] 

\begin{figure}[ht]
\vskip 0.2in
\begin{center}
\centerline{\includegraphics[width=\columnwidth]{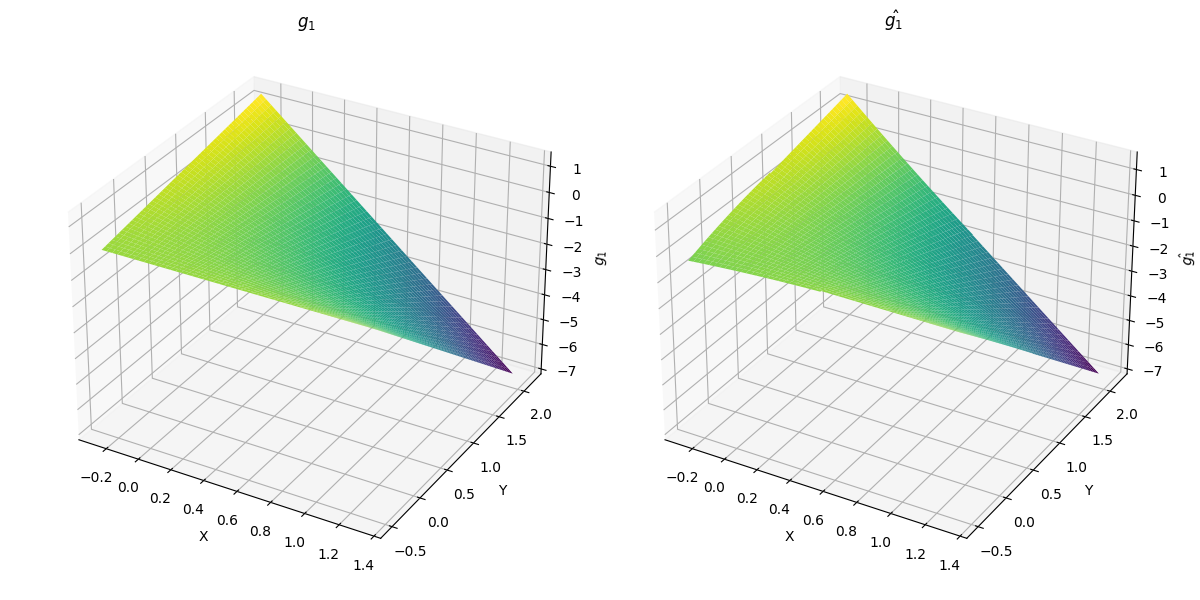}}
\caption{Comparison of $\evg_1$ (left) and $\hat{\evg_1}$ (right) for  H\'{e}non-Heiles system~\eqref{eq:henon_heiles_system}.}
\label{fig:HH g_1}
\end{center}
\vskip -0.2in
\end{figure}

\begin{figure}[ht]
\vskip 0.2in
\begin{center}
\centerline{\includegraphics[width=\columnwidth]{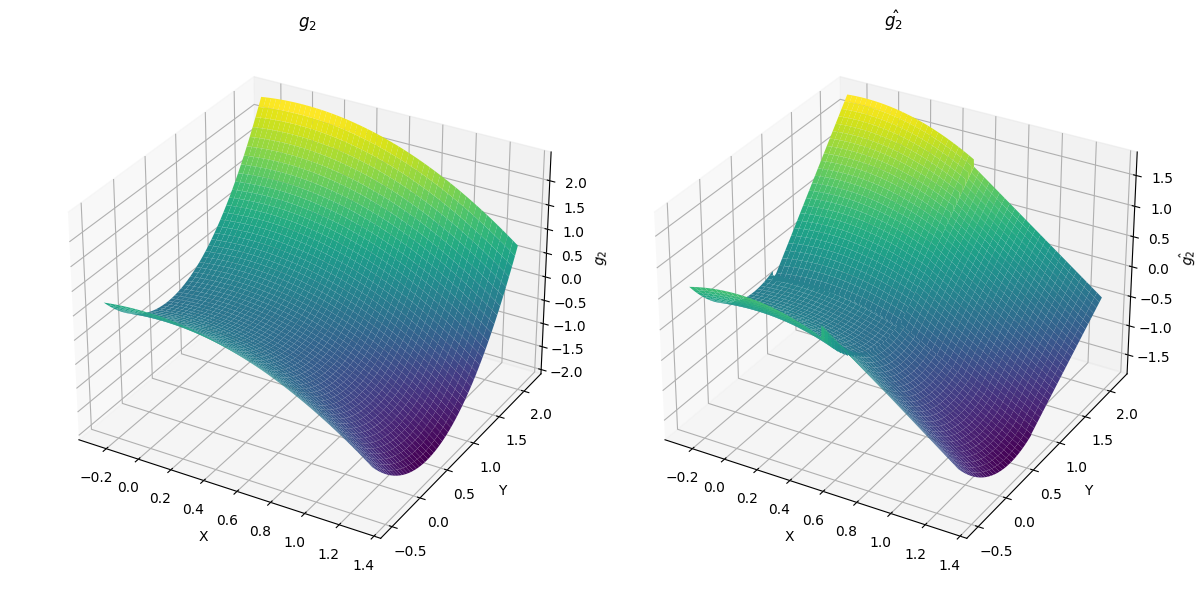}}
\caption{Comparison of $\evg_2$ (left) and $\hat{\evg_2}$ (right) for  H\'{e}non-Heiles system~\eqref{eq:henon_heiles_system}.}
\label{fig:HH g_2}
\end{center}
\vskip -0.2in
\end{figure}

In simulation of trajectories of H\'{e}non-Heiles System, we set $\lambda = 1$. Figures \ref{fig:HH g_1} and \ref{fig:HH g_2} display the comparison of true and estimated drift functions for each component of $\rvg$. Moreover, Table \ref{tab:HHS} presents the performance measures of drift function estimation. The estimation of noise structure of H\'{e}non-Heiles System is displayed in Table \ref{tab:HHS sigma}.

\begin{table}
    \centering
\caption{H\'{e}non-Heiles system drift estimation summary}
\label{tab:HHS}
    \begin{tabular}{|c|c|} \hline 
         Relative $L^2(\rho)$ Error& 0.106\\ \hline 
         Relative Trajectory Error& $0.076$ $\pm$ $0.053$\\ \hline 
         Wasserstein Distance at $t = 0.25$& 0.0529\\ \hline
 Wasserstein Distance at $t = 0.5$&0.0694\\\hline
 Wasserstein Distance at $t = 1$&0.0904\\\hline
    \end{tabular}

\end{table}

\begin{table}
    \centering
\caption{H\'{e}non-Heiles system diffusion estimation}
\label{tab:HHS sigma}
    \begin{tabular}{|c|c|} \hline 
           True $\sigma_1$& Estimated $\hat{\sigma_1}$\\ \hline 
           0.0700& 0.0700\\ \hline \hline
 True $\sigma_2$&Estimated $\hat{\sigma_2}$\\ \hline 
 0.0500&0.0500\\ \hline
    \end{tabular} 
\end{table}

\subsection{Stochastic Cucker-Smale system}\label{subsec:collective}
\begin{figure}[ht]
\vskip 0.2in
\begin{center}
\centerline{\includegraphics[width=\columnwidth]{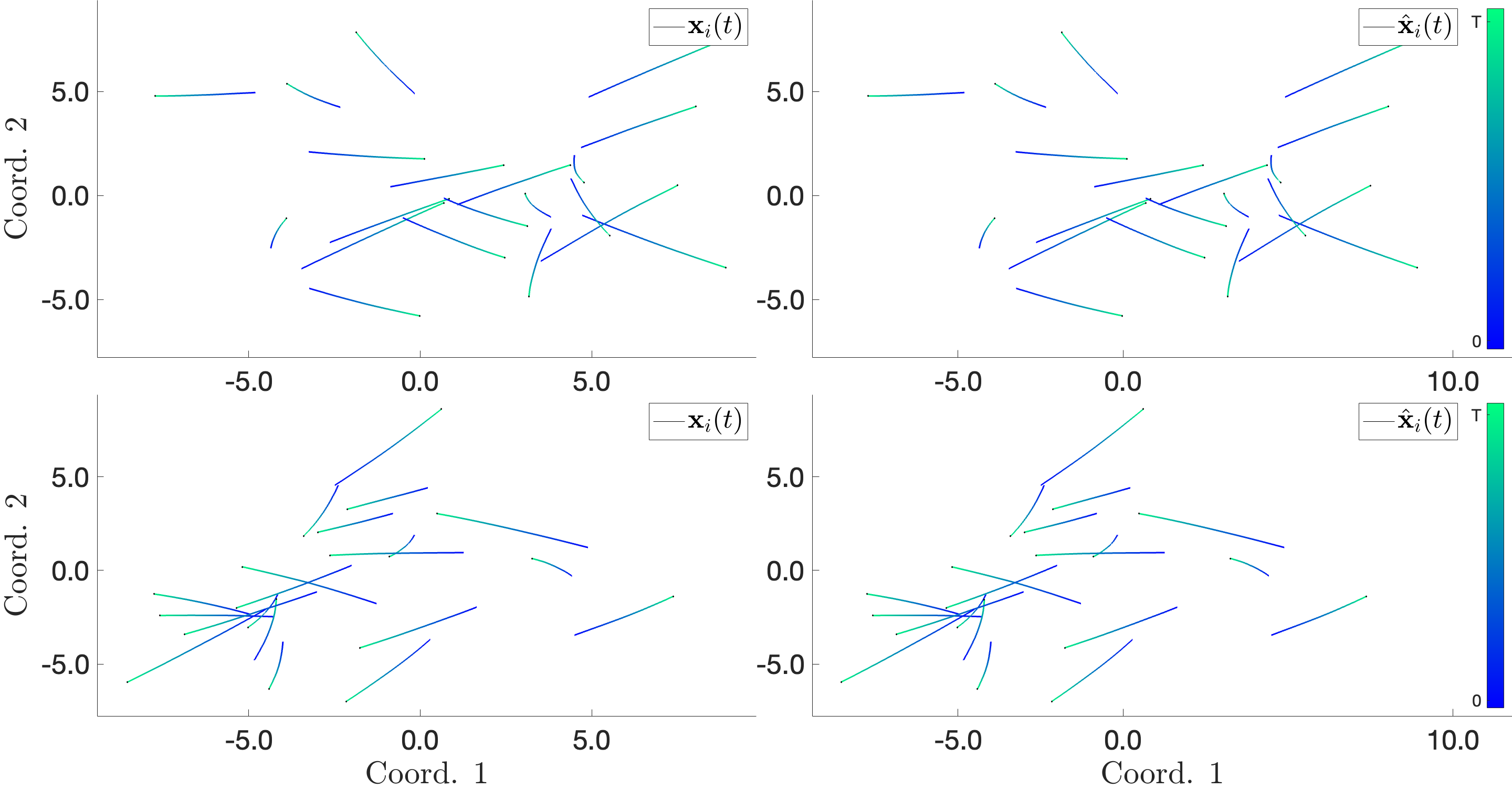}}
\caption{Comparison of $\hat\rvx_i(t)$ vs $\rvx_i(t)$ in the Stochastic Cucker-Smale system}
\label{fig:sto_cs_trjas}
\end{center}
\vskip -0.2in
\end{figure}

We are interested in a particular family of interacting agent systems, namely collective dynamical systems, which can be considered as a high-dimensional mSDE. For example, we consider the stochastic Cucker-Smale flocking dynamics for a system of $N$ agents as follows:
\begin{equation}\label{eq:sto_cs}
\begin{aligned}
\dif\rvx_i &= \rvv_i\dt, \\
\dif\rvv_i &= \big(\frac{1}{N}\sum_{j = 1, j \neq i}^N\phi^A(\norm{\rvx_j - \rvx_i})(\rvv_j - \rvv_i)\big)\dt \\
&\quad + \sigma(\rvv_i)\dif\rvw_i^{\rvv},
\end{aligned}
\end{equation}
for $i = 1, \cdots. N$.  Here $\rvx_i, \rvv_i \in \R^d$ is the position/velocity of the $i^{th}$ bird respectively, $\rvw_i^{\rvv}$ is the standard Brownian motion, the function $\phi^a: \R^+ \rightarrow \R$ is known as an alignment based interaction function which governs the force that the $j^{th}$ agent exerts on the $i^{th}$ agent, and the noise $\sigma: \R^{d} \rightarrow \R$. When we let
\[
\rvx = \begin{bmatrix} \rvx_1 \\ \vdots \\ \rvx_N \end{bmatrix} \in \R^{D = Nd} \quad \text{and} \quad \rvy = \begin{bmatrix} \rvv_1 \\ \vdots \\ \rvv_N \end{bmatrix} \in \R^{D},
\]
moreover, we define the drift term as
\[
\rvg(\rvx, \rvy) = \begin{bmatrix}\frac{1}{N}\sum_{j = 2}^N\phi(\norm{\rvx_j - \rvx_1})(\rvv_j - \rvv_1) \\ \vdots \\ \frac{1}{N}\sum_{j = 1}^{N - 1}\phi(\norm{\rvx_j - \rvx_N})(\rvv_j - \rvv_N) \end{bmatrix},
\]
and the noise term as $\msigma^{\rvy} = \msigma^{\rvy}(\rvy)$ is defined as
\[
\msigma^{\rvy} = \begin{bmatrix}\sigma(\rvv_1)\mI_{d\times d} & \rvzero_{d\times d} & \cdots & \rvzero_{d\times d} \\\rvzero_{d\times d} & \sigma(\rvv_2)\mI_{d\times d} & \cdots & \rvzero_{d\times d} \\\vdots & \vdots & \ddots & \vdots \\ \rvzero_{d\times d} & \cdots & \rvzero_{d\times d} & \sigma(\rvv_N)\mI_{d\times d}  \end{bmatrix}
\]
Notice that each $\sigma(\rvv_i)$ is a scalar.  Then we can obtain the original formula as introduced in~\eqref{eq:mix_general}.  However the system now becomes extremely high-dimensional as $D = Nd$. But by combining the techniques in \cite{lu2021learning, GCZ2024} and using the losses introduced in Section \ref{sec:learning_framework}, we are able to obtain the following results.  The simulation presented here considers $N = 20$ agents, with $\phi^A = \frac{1}{(1 + r^2)^{0.25}}$, and $\sigma = 0.1$.
\begin{figure}[ht]
\vskip 0.2in
\begin{center}
\centerline{\includegraphics[width=\columnwidth]{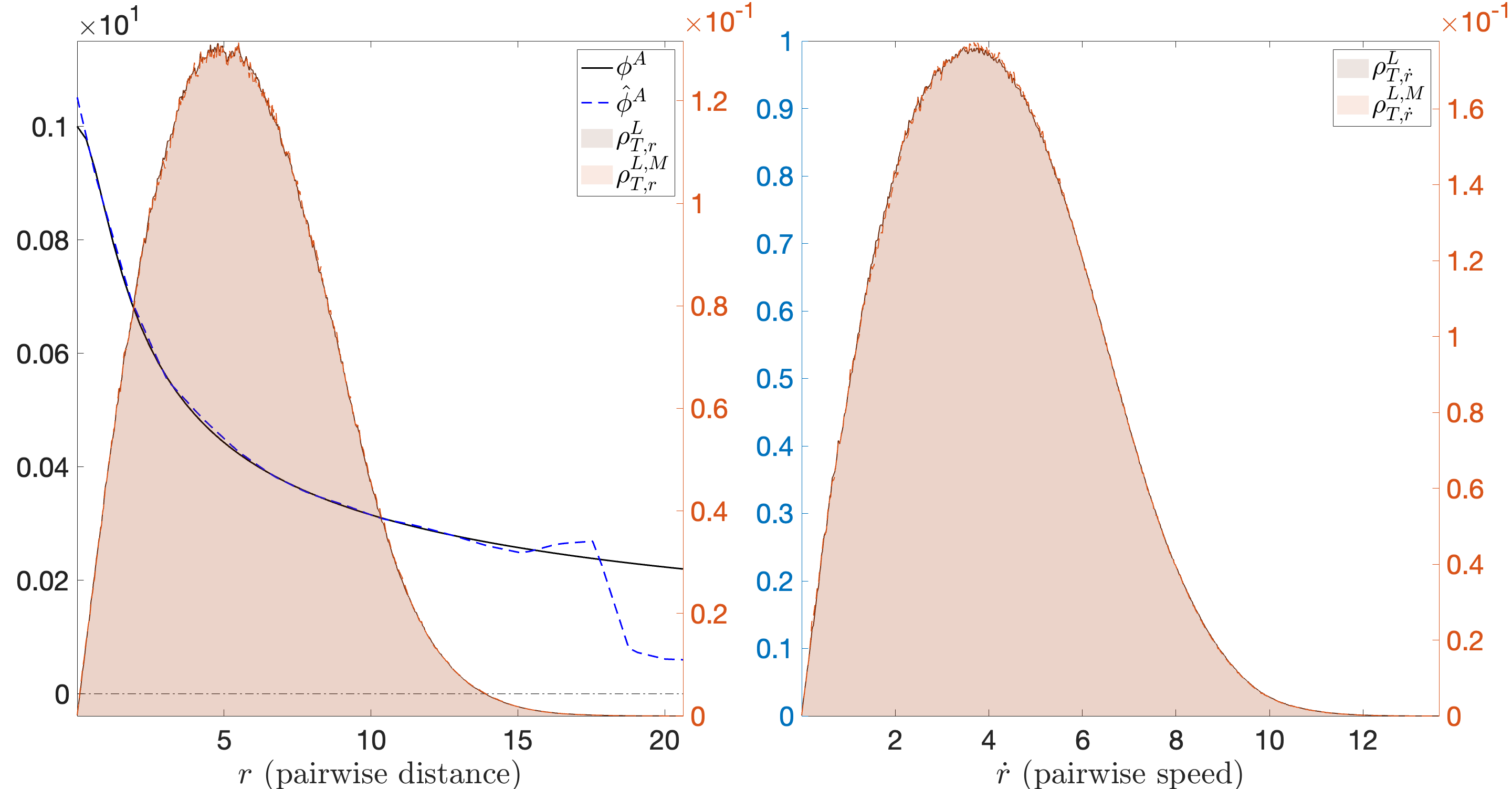}}
\caption{Comparison of $\hat\phi^A$ vs $\phi^A$ in the stochastic Cucker-Smale system}
\label{fig:sto_cs_phi}
\end{center}
\vskip -0.2in
\end{figure}
Figure \ref{fig:sto_cs_phi} shows the estimation of $\phi^A$ and Figure \ref{fig:sto_cs_trjas} shows the estimation of $\rvx_i(t)$.
\section{Conclusions and future work}\label{sec:conc_future}
We have demonstrated that our learning framework for SDEs with structured noised (denoted as mSDEs) for various different example systems, including the van der Pol oscillator, a modified Vicsek model of active matter, the H\'{e}non-Heiles system chaotic Hamiltonian system, and a stochastic Cucker-Smale alignment model.  Our results suggest high-fidelity learning with mechanistic estimators.  The results presented here are preliminary, and the true application to high-dimensional systems with proper dimension reduction methods is currently being developed.  Specifically, we are currently deriving methods to discover the feature maps $\mxi_{\rvf}$ and $\mxi_{\rvg}$ together with the drifts $\rvf$ and $\rvg$ in~\eqref{eq:mix_general}.  We believe that the learning of the proper feature maps will significantly reduce the dimension of the desired drift and diffusion terms, and hence will improve learning accuracy.
\section*{Impact Statement}



This paper presents work whose goal is to combine the current state-of-the-art machine learning methods with applications in physics, biology, chemistry, and finance.  There are many potential societal consequences of our work, including its utilization as a basis for data-driven modeling of social behaviors such as crime modeling.  Furthermore, our methods can be applied broadly, including in the study of animal conservation, pedestrian dynamics for safe-city design, crowd dynamics for emergency evacuation, and the network dynamics of autonomous vehicles.
\bibliography{learning_bib}
\bibliographystyle{icml2025}
\end{document}